\documentclass[11pt]{article}

\usepackage[final]{acl}
\usepackage{balance}
\usepackage{times}
\usepackage{latexsym}
\usepackage[T1]{fontenc}
\usepackage[utf8]{inputenc}
\usepackage{microtype}
\usepackage{inconsolata}
\usepackage{fontawesome5}

\usepackage{amsmath} 

\newcommand{\res}[2]{$#1_{_{\pm #2}}$}

\usepackage{hyperref} 
\usepackage{url}
\usepackage{tikz}
\usepackage{nicematrix}
\usepackage{booktabs}
\usepackage{makecell}
\usepackage{multirow}
\usepackage{arydshln}
\usepackage{enumitem}
\usepackage{tabularx}
\usepackage{pifont}
\usepackage[breakable]{tcolorbox}
\usepackage[table]{xcolor} 
\usepackage{amssymb}
\usepackage{afterpage}
\usepackage{wrapfig}
\usepackage{graphicx}   
\usepackage{subcaption} 

\newtcolorbox{takeawaybox}[1][]{
    breakable,
    colback=black!5,
    colframe=black!70,
    boxrule=1pt,
    arc=2mm,
    fonttitle=\bfseries,
    #1
}

\newtcolorbox{Hypothesis}[1][]{
    breakable,
    colback=black!5,
    colframe=black!70,
    boxrule=1pt,
    arc=2mm,
    fonttitle=\bfseries,
    #1
}

\newtcolorbox{promptbox}[1][]{
    breakable,
    colback=gray!10, 
    colframe=gray!80,
    boxrule=1pt,
    arc=2mm,
    fonttitle=\bfseries\small, 
    #1
}

\hypersetup{
    colorlinks=true,
    linkcolor=black,
    citecolor=cyan,
    filecolor=magenta,      
    urlcolor=magenta,
}

\newtcolorbox{premisebox}{
  colback=gray!10,        
  colframe=gray!60,       
  fonttitle=\bfseries,      
  title=(Counterfactual Premise), 
  boxrule=1pt,            
  arc=2mm,                
}

\newtcolorbox{comparisonbox}{
  breakable,              
  colframe=black!15,      
  boxrule=0.5pt,          
  arc=0mm,                
  toptitle=5mm,           
  bottomtitle=5mm,       
  fonttitle=\bfseries,      
  colback=blue!5,         
  sidebyside,             
  lower separated=true,   
  sidebyside align=top,   
}

\title{Rethinking RL Evaluation: Can Benchmarks Truly Reveal Failures of RL Methods?}

\author{
Zihan Chen\thanks{Equal contribution. \textsuperscript{$\dagger$} Corresponding authors. Correspondence to: Zenghui Ding \href{mailto:dingzenghui@iim.ac.cn}{\texttt{<dingzenghui@iim.ac.cn>}} and Cho-Jui Hsieh \href{mailto:chohsieh@cs.ucla.edu}{\texttt{<chohsieh@cs.ucla.edu>}}.}\textsuperscript{1,2},
Yiming Zhang\footnotemark[1]\textsuperscript{1,2},
Hengguang Zhou\textsuperscript{3},
\textbf{Zenghui Ding}\textsuperscript{$\dagger$1},\\
\textbf{Yining Sun}\textsuperscript{1},
\textbf{Cho-Jui Hsieh}\textsuperscript{$\dagger$3,4} \\
\textsuperscript{1}HFIPS, Chinese Academy of Sciences \quad
\textsuperscript{2}University of Science and Technology of China \\
\textsuperscript{3}University of California, Los Angeles \quad
\textsuperscript{4}Arena \\
\vspace{0.35em}
{\small \faGithub\ \textbf{Project:} \href{https://rethinkingrl.github.io/RL-GAP.github.io/}{RL-GAP.github.io}}
}

\begin{document}
\maketitle

\begin{abstract}
Current benchmarks are inadequate for evaluating progress in reinforcement learning (RL) for large language models (LLMs). Despite recent benchmark gains reported for RL, we find that training on these benchmarks' training sets achieves nearly the same performance as training directly on the test sets, suggesting that the benchmarks cannot reliably separate further progress. To study this phenomenon, we introduce a diagnostic suite and the Oracle Performance Gap (OPG) metric that quantifies the performance difference between training on the train split versus the test split of a benchmark. We further analyze this phenomenon with stress tests and find that, despite strong benchmark scores, existing RL methods struggle to generalize across distribution shifts, varying levels of difficulty, and counterfactual scenarios: shortcomings that current benchmarks fail to reveal.We conclude that current benchmarks are insufficient for evaluating generalization and propose three core principles for designing more faithful benchmarks: sufficient difficulty, balanced evaluation, and distributional robustness.
\end{abstract}

\section{INTRODUCTION}
Reinforcement Learning (RL) has emerged as a powerful paradigm for post-training Large Language Models (LLMs), significantly enhancing their capabilities on complex, multi-step reasoning tasks \citep{10.5555/3600270.3602281}. Methods based on Reinforcement Learning from Human Feedback (RLHF) and Direct Preference Optimization (DPO)~\citep{rafailov_direct_2023} have become standard practice for aligning LLMs. These paradigms are often powered by foundational algorithms like Proximal Policy Optimization (PPO)~\citep{schulman2017proximal}, with state-of-the-art variants such as Group Relative Policy Optimization (GRPO)~\citep{shao_deepseekmath_2024} pushing models to achieve remarkable performance on benchmarks like GSM8K~\citep{cobbe2021gsm8k} and MATH~\citep{hendrycks_measuring_2021}. These successes, marked by state-of-the-art results (\citealp{NEURIPS2022_18abbeef}; \citealp{lightman2023let}),
\begin{figure*}[t]
    \centering
    \includegraphics[width=1.0\textwidth]{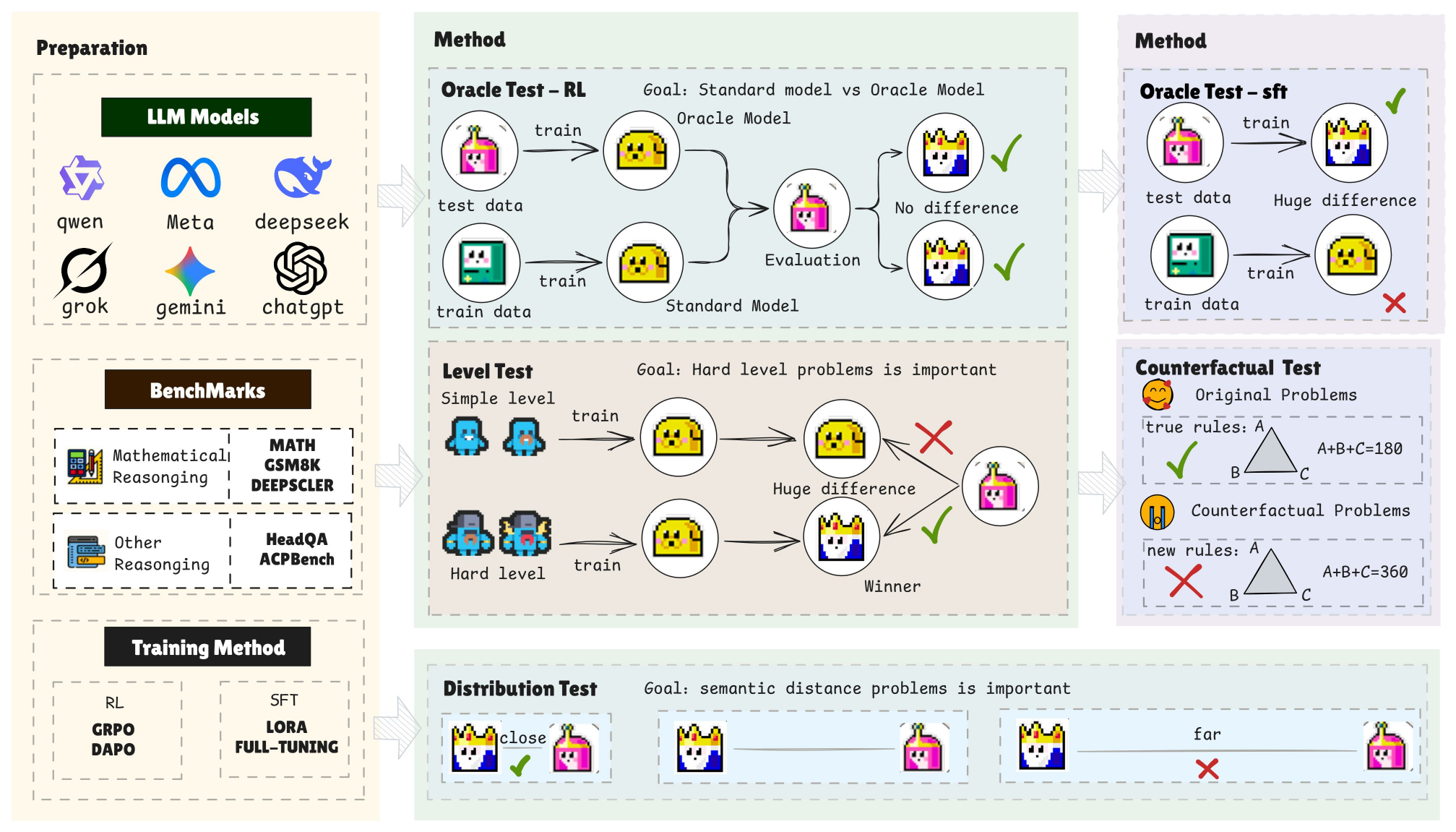}

    \caption{
        \textit{Overview of our empirical framework. }The workflow begins by diagnosing benchmark flaws with novel metrics to uncover a core symptom: a vanishing generalization gap. It then proceeds through a suite of stress tests that reveal the brittle, shortcut-based nature of the learned skills, culminating in a new set of principles for more robust evaluation.
    }
    \label{fig:overview_flowchart}
\end{figure*}

suggest that RL-based alignment is a key pathway toward robust reasoning systems. Yet, a critical question remains: do current benchmarks meaningfully assess generalization? We find that the traditional assumption---that performance on unseen data measures generalization---may be insufficient for RL, as models trained on training splits perform nearly identically to those trained directly on test splits. This suggests that ``unseen-ness'' may no longer be a sufficiently discriminative criterion, calling for evaluations that go beyond disjoint splits to reveal deeper weaknesses.

To systematically investigate this phenomenon, we introduce an empirical evaluation framework for assessing whether the conventional train--test split remains a meaningful indicator of generalization for RL-trained models~\citep{yu2025frame}. At the core of this framework, we define the Oracle Performance Gap (OPG) as a primary diagnostic measuring the performance difference between train-split-optimized and test-split-optimized models on the same benchmark. We further complement our analysis with targeted stress tests that probe whether high benchmark scores continue to correlate with robust generalization. These tests examine performance under variations in difficulty, question type, and counterfactual settings, revealing discrepancies between saturated benchmark performance and generalization behavior. Importantly, the observed trends are consistent with the Oracle Performance Gap analysis, jointly suggesting that near-ceiling benchmark scores alone may be insufficient to reliably assess generalization. These findings are then used to motivate three principles for benchmark design. Overall, our contributions are:
\begin{itemize}[leftmargin=*, topsep=3pt, itemsep=2pt]
    \item[\ding{68}] \textbf{Illusion of Capability.} We present quantitative analyses suggesting that high performance on prevailing benchmarks does not necessarily correspond to robust generalization. Using the Oracle Performance Gap and aligned stress tests (including distributional and counterfactual evaluations), we identify structural limitations under which benchmarks assign strong scores despite observable discrepancies in generalization behavior. These findings highlight limitations in the reliability of current benchmarks scores as indicators of reasoning ability.
    
    \item[\ding{68}] \textbf{Novel Diagnostic Framework.} We introduce a new diagnostic framework, including the OPG and a set of evaluations (difficulty, distributional, and counterfactual), to systematically probe and quantify the extent to which benchmark scores remain informative about the generalization capability of RL models, rather than merely reflecting benchmark-specific fitting.

    \item[\ding{68}] \textbf{Actionable Design Principles.} Based on our findings, we propose a set of actionable principles for designing next-generation benchmarks that can more robustly evaluate an agent's true, transferable reasoning abilities under more challenging and realistic evaluation settings.

\end{itemize}
\section{Diagnosing Generalization via OPG}
\label{sec:Benchmark Flaws}

The standard approach to evaluating LLM reasoning is to measure performance on a held-out test set, under the assumption that success on unseen data reflects generalization. To examine whether this assumption still holds for RL-based methods, we introduce a diagnostic framework that tests whether “unseen-ness”, the common practice of relying on disjoint train/test splits, continues to provide a valid measure of generalization. Our framework compares RL models trained on training split with the Oracle model trained directly on the test split and finds that their performance is nearly identical, indicating that test-set "unseen-ness" alone has ceased to be a diagnostic signal of generalization.

\subsection{Analysis Framework}

\subsubsection{Oracle Performance Gap (OPG)}

\begin{table*}[t]
\caption{\textit{Benchmark Limitation Illustrated by Qwen2.5 Model Performance.} The table is reorganized by benchmark, comparing performance across 3B and 7B model scales.}
\label{tab:qwen-performance-subscript}
\centering
\small
\setlength{\tabcolsep}{3.5pt}
\begin{tabular*}{\textwidth}{
    l l @{\extracolsep{\fill}}
    ccccc cc
}
\toprule
\multirow{2}{*}{\textbf{Benchmark}} & \multirow{2}{*}{\makecell[l]{\textbf{Model}\\\textbf{Size}}} & \multicolumn{4}{c}{\textbf{RL on Train Set Subsets (\%)}} & \multirow{2}{*}{\makecell{\textbf{RL Oracle} \\ \textbf{($M_{RL,test}$)}} } & \multirow{2}{*}{\makecell{\textbf{Baseline} \\ \textbf{($M_{base}$)}} } & \multirow{2}{*}{\makecell{\textbf{OPG} \\ \textbf{(\%)}} } \\
\cmidrule(r){3-6} 
& & \textbf{10\%} & \textbf{20\%} & \textbf{50\%} & \textbf{100\%} & & & \\
\midrule
\multirow{2}{*}{MATH} 
& 3B & \res{63.88}{1.04} & \res{65.18}{0.84} & \res{64.84}{1.25} & \res{64.62}{0.98} & \res{64.62}{1.11} & 62.20 & 0.00 \\
& 7B & \res{73.64}{0.48} & \res{73.28}{0.68} & \res{73.04}{0.91} & \res{74.04}{0.39} & \res{74.00}{0.68} & 68.80 & -0.05 \\
\midrule 
\multirow{2}{*}{GSM8K} 
& 3B & \res{82.95}{0.38} & \res{83.93}{0.47} & \res{86.93}{0.34} & \res{87.04}{0.49} & \res{87.98}{0.35} & 83.02 & 1.07 \\
& 7B & \res{88.76}{0.47} & \res{89.58}{0.44} & \res{91.14}{0.30} & \res{91.72}{0.31} & \res{91.87}{0.31} & 88.40 & 0.16 \\
\midrule 
\multirow{2}{*}{DeepScaler} 
& 3B & \res{34.12}{0.94} & \res{32.68}{1.02} & \res{35.75}{0.96} & \res{35.22}{0.89} & \res{34.95}{0.91} & 33.38 & -0.77 \\
& 7B & \res{42.05}{0.84} & \res{42.09}{0.57} & \res{42.84}{0.73} & \res{42.36}{0.77} & \res{42.64}{0.75} & 35.70 & 0.66 \\
\midrule 
\multirow{2}{*}{HeadQA} 
& 3B & \res{62.98}{0.77} & \res{65.90}{1.19} & \res{67.16}{0.79} & \res{67.24}{0.59} & \res{67.57}{0.70} & 54.96 & 0.49 \\
& 7B & \res{72.94}{0.90} & \res{72.90}{0.79} & \res{74.39}{0.60} & \res{75.20}{0.66} & \res{75.60}{0.50} & 52.24 & 0.53 \\
\bottomrule
\end{tabular*}
\end{table*}
We introduce the Oracle Performance Gap (OPG) as a diagnostic metric to audit the validity of a benchmark. Formally, let $P(M, \mathcal{D})$ denote the pass@1 accuracy of a model $M$ on dataset $\mathcal{D}$ using algorithm $\mathcal{A} \in \{\text{SFT, RL}\}$. We define OPG as:
\begin{equation}
    \text{OPG}_{\mathcal{A}} \triangleq \frac{P(M_{\mathcal{A},\text{test}}, \mathcal{D}_{\text{test}}) - P(M_{\mathcal{A},\text{train}}, \mathcal{D}_{\text{test}})}{P(M_{\mathcal{A},\text{test}}, \mathcal{D}_{\text{test}})}.
\end{equation}
Here, $M_{\mathcal{A},\text{train}}$ is the standard model and $M_{\mathcal{A},\text{test}}$ is the ``Oracle'' model fine-tuned explicitly on the test set. Unlike standard generalization gaps (which measure overfitting), OPG measures the \textit{discriminative power} of the test set by quantifying the performance deficit against this Oracle baseline.

With OPG, we establish an upper bound for performance via memorization and assess whether the benchmark effectively challenges algorithm $\mathcal{A}$. We distinguish two outcomes:
\begin{enumerate}[leftmargin=*, noitemsep, topsep=1pt, partopsep=0pt]
    \item \textbf{Effective Generalization ($\text{OPG}_{\mathcal{A}} \gg 0$):} A significant gap suggests that the benchmark poses a meaningful challenge, as the test set contains specific patterns or difficulties that cannot be trivially inferred from the training set.
    \item \textbf{Weak Discriminative Signal ($\text{OPG}_{\mathcal{A}} \lesssim 0$):} A negligible gap indicates structural redundancy. It suggests the test set fails to differentiate between true generalization and simple pattern matching, as ``seeing'' the test data offers no performance advantage.
\end{enumerate}

\label{sec:oracle_analysis}
\subsubsection{Experiment Setup}
\label{sec:Experiment_Setup}
Our analysis spans four benchmarks: MATH, GSM8K, HeadQA, and DeepScaler. We use two base models from the \texttt{Qwen} family, \texttt{Qwen2.5-3B-Instruct} and \texttt{Qwen2.5-7B-Instruct}. To systematically isolate the effects of the fine-tuning paradigm and data distribution, we create and compare a suite of six model variants for each base model:
\begin{itemize}[leftmargin=*, topsep=3pt, itemsep=2pt]
    \item \textbf{Baseline ($M_{base}$):} The original instruction-tuned model without any additional fine-tuning.
    \item \textbf{Standard SFT ($M_{SFT,train}$):} The base model fine-tuned on the official training set containing only standard question-answer pairs.
    \item \textbf{SFT with CoT ($M_{SFT,formatted}$):} The base model fine-tuned on a formatted training set that includes detailed, teacher-generated chain-of-thought (CoT) reasoning steps.
    \item \textbf{RL on Train Set ($M_{RL,train}$):} The base model fine-tuned on the official training set using GRPO, a state-of-the-art RL algorithm.
    \item \textbf{SFT Oracle ($M_{SFT,test}$):} The base model fine-tuned directly on the \textit{test set} using SFT. This serves as a practical upper bound for SFT performance on the test distribution.
    \item \textbf{RL Oracle ($M_{RL,test}$):} The base model fine-tuned directly on the \textit{test set} using the same GRPO setup. This provides an upper bound for the RL agent's ability to exploit the test set.
\end{itemize}

All models are evaluated on the official test sets using pass@1 accuracy. To ensure statistical rigor, we rigorously quantified performance stability by conducting 10 independent evaluation runs via sampling for each reported metric. Accordingly, we report the \textbf{Mean $\pm$ 95\% Confidence Interval (CI)}, providing a robust measure of reliability. Full implementation details, hyperparameters, and evaluation protocols are provided in Appendix~\ref{sec:exp_setup}.

\subsection{Result}

\begin{table*}[t]
    \centering
    \caption{\textit{SFT Performance Reveals the Expected Generalization Gap.} This table presents the SFT results organized by benchmark, with a direct comparison between the 3B and 7B model scales. All values are pass@1 accuracy.}
    \label{tab:sft-performance-booktabs-recalculated}
    \begin{tabular*}{\textwidth}{
        l l @{\extracolsep{\fill}} 
        ccc c
    }
        \toprule
        \multirow{2}{*}{\textbf{Benchmark}} & \multirow{2}{*}{\makecell[l]{\textbf{Model}\\\textbf{Size}}} & \multicolumn{3}{c}{\textbf{SFT Performance Metrics (\%)}} & \multirow{2}{*}{\makecell{\textbf{OPG} \\ \textbf{(\%)}}} \\
        \cmidrule(r){3-5}
        & & \textbf{$M_{SFT,train}$} & \textbf{$M_{SFT,test}$} & \textbf{$M_{SFT,formatted}$} & \\
        \midrule
        
        \multirow{2}{*}{MATH} 
        & 3B & \res{17.20}{0.85} & \res{40.00}{1.12} & \res{31.02}{0.95} & 22.45 \\
        & 7B & \res{23.60}{0.78} & \res{64.20}{0.65} & \res{42.00}{0.82} & 34.58 \\
        \midrule

        \multirow{2}{*}{GSM8K} 
        & 3B & \res{16.83}{0.45} & \res{68.05}{0.41} & \res{64.82}{0.39} & 4.75 \\
        & 7B & \res{19.71}{0.42} & \res{79.04}{0.33} & \res{75.36}{0.35} & 4.66 \\
        \midrule

        \multirow{2}{*}{DeepScaler} 
        & 3B & \res{8.51}{0.92} & \res{27.03}{0.98} & \res{22.57}{0.94} & 16.50 \\
        & 7B & \res{12.57}{0.85} & \res{36.76}{0.76} & \res{23.51}{0.80} & 36.04 \\
        \midrule

        \multirow{2}{*}{HeadQA} 
        & 3B & \res{10.45}{0.68} & \res{54.96}{0.75} & \res{41.22}{0.82} & 25.00 \\
        & 7B & \res{12.80}{0.62} & \res{52.24}{0.66} & \res{35.52}{0.74} & 32.01 \\

        \bottomrule
    \end{tabular*}
\end{table*}

\paragraph{Finding 1: The Vanishing Generalization Gap Suggests Unseen-ness is an Insufficient Criterion.}

The OPG analysis reveals a stark contrast between SFT and RL paradigms, as detailed in Tables~\ref{tab:qwen-performance-subscript} and~\ref{tab:sft-performance-booktabs-recalculated}. While SFT models exhibit a large and expected OPG in a challenging generalization setting, this gap collapses to near-zero for RL-trained models. To rule out the concern that this is caused by data leakage in the base model, we verified that both our fine-tuned models significantly outperform the untrained baseline, confirming that this result reflects standard training behavior. To further assess if our conclusion is general, we also evaluated additional RL algorithms such as DAPO, as well as alternative architectures, domains, and inference settings; the corresponding results are presented in Appendix~\ref{app:opg_robustness}.

Furthermore, to account for potential effect of the test set being much smaller than the training set, we trained RL models on various subsets of the training data. The OPG remained consistently low across all sizes, suggesting our conclusion is robust to the effects of data quantity. Together, these results suggest that the classical assumption—that performance on “unseen” test data is a sufficient measure of generalization—no longer holds for RL. This suggests that current benchmarks may no longer meaningfully assess future progress in RL generalization.

\begin{takeawaybox}
\textbf{Takeaway~\ding{202}.}
Our OPG analysis reveals that RL models trained on the training split perform nearly identically to Oracle models trained directly on the test split. This performance convergence suggests that existing datasets likely suffer from structural redundancy, rendering the traditional criterion of test-set “unseen-ness” insufficient. Consequently, simple train-test splitting may no longer provide a sufficiently discriminative measure of true generalization for RL models.
\end{takeawaybox}
\paragraph{Robustness beyond the primary setting.}
We additionally evaluated whether the near-zero OPG trend persists beyond the primary Qwen2.5-based mathematical setting. We found similarly small OPG values across an alternative RL algorithm (DAPO; \citealp{yu_dapo_2025}), an additional open-weight model family (Llama-3-8B; \citealp{grattafiori_llama_2024}), and non-mathematical reasoning domains including HotpotQA \citep{yang2018hotpotqa}, MedQA \citep{jin2021disease}, and LogiQA \citep{liu2020logiqa}, as well as under varied KL coefficients and decoding parameters. These results provide further support that the observed vanishing-gap phenomenon is not specific to a single algorithm, architecture, domain, or inference setting. Detailed results are provided in Appendix~\ref{app:opg_robustness}.

\section{Benchmark Principles}

Section~\ref{sec:Benchmark Flaws} showed, via the Oracle Performance Gap, that standard benchmark evaluations may fail to reliably reflect generalization. In this section, we complement that analysis by evaluating models under more revealing evaluation settings that make specific dimensions of generalization explicit and provide insight into directions for improving benchmark design. By examining performance across difficulty levels, distributional shifts, and counterfactual conditions, we observe patterns consistent with the OPG findings: models achieving near-saturated benchmark performance can nonetheless exhibit pronounced performance degradation under these settings. Together, these analyses expose three structural limitations in existing benchmarks and motivate three corresponding principles for benchmark design: Cross-Difficulty Generalization, Distributional Robustness, and counterfactual Reasoning.

\label{sec:h2-deconstruction}
\begin{figure*}[tbp] 
    \centering 
    \includegraphics[width=\textwidth]{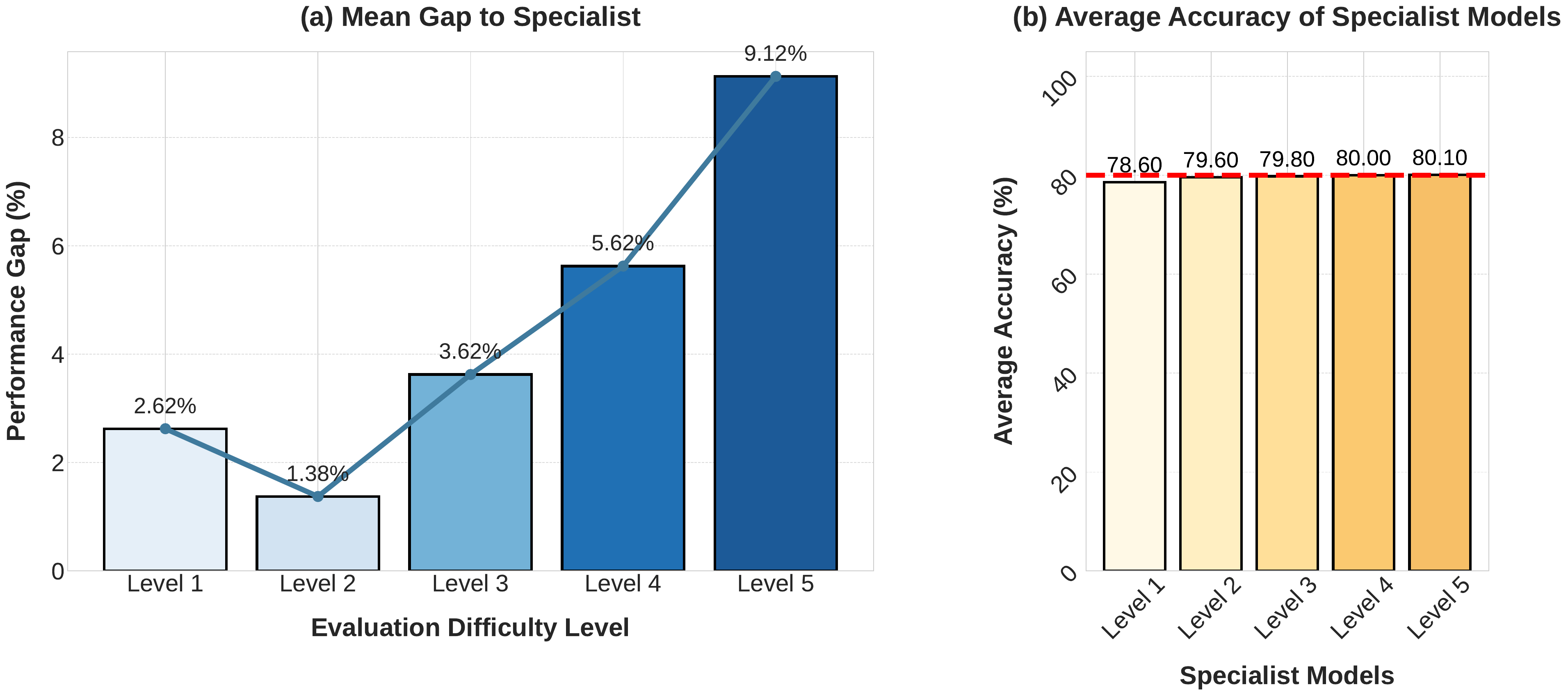}
    \caption{
    \textit{The Illusion of Average Performance.} 
    \textbf{(a)} The mean performance gap between the best (specialist) model and the average of all other models widens dramatically as task difficulty increases. 
    \textbf{(b)} Surprisingly, the average scores of these specialists (calculated across all five difficulty partitions) are nearly identical. This contrast illustrates how a difficulty-agnostic evaluation can mask substantial differences in generalization capability. Full performance data is provided in Appendix~\ref{app:c_result}.
}
    \label{fig:final_comparison} 
\end{figure*}
\subsection{The Difficulty Test}
\label{sec:difficulty-stratified-analysis}
One plausible factor contributing to the failure in evaluation discussed in Section~\ref{sec:Benchmark Flaws} is that typical train--test splits do not account for variation in sample difficulty, instead summarizing performance by aggregating test instances across difficulty levels. This form of aggregation can obscure systematic differences in generalization behavior, particularly when failures are concentrated on more challenging cases and are diluted by easier ones. To make this dimension explicit, we stratify benchmarks by task difficulty and analyze performance beyond aggregate scores.
\subsubsection{The Paradox of Average Scores}
\label{sec:average_score}

\textbf{Setup.}
We conduct a cross-difficulty analysis by training five specialist models ($M_{L_i}$), each fine-tuned on a single difficulty partition of our constructed \textbf{MATH dataset} ($\mathcal{D}_{\text{train}}^{L_i}$, see Appendix~\ref{app:difficulty_annotation} for the partition protocol). Each specialist is then evaluated on all five training sets partitions, where a model's performance on its own training data serves as an oracle's performance benchmark against which its true generalization to unseen partitions is measured.

We observe that RL models exhibit asymmetric generalization: models trained on harder levels transfer well to easier ones, while those trained on easier levels struggle to generalize to harder tasks (Figure~\ref{fig:cross-difficulty}). Yet, when computing a single average score—mirroring the ''vanishing generalization gap''—these models achieve nearly identical results (Figure~\ref{fig:final_comparison}(b)). This indicates that the OPG did not truly vanish; instead, it was concealed by difficulty-agnostic averaging. Thus, standard averaging masks meaningful differences, creating a misleading impression of equal capability that only stratified evaluation can reveal.
\vspace{-2mm}
\paragraph{Finding 2: A difficulty-aware train--test split is an effective setting for evaluating generalization.}
Our cross-difficulty evaluation confirms failure modes concealed by average scores, suggesting that difficulty-aware partitioning is a superior paradigm for evaluation. More importantly, it reveals differences in transfer behavior that are invisible under standard aggregate metrics. This is supported by two key findings:
\begin{itemize}[noitemsep,topsep=1pt,leftmargin=*]
    \item \textbf{Masking Effect Confirmed:} Figure~\ref{fig:final_comparison}(b) reveals that significant capability variations are completely masked by final average scores. Although the specialist models differ substantially in their cross-difficulty behavior, the final average scores across them remain nearly identical.
    
    \item \textbf{Oracle Gap Re-emerges at the Micro-Level:} Contrary to the global near-zero OPG, micro-level analysis reveals a persistent gap. As shown in Figure~\ref{fig:final_comparison}(a), the gap against the specialist oracle ($M_{L_j}$) is non-zero; instead, it reappears and widens as task complexity increases. This shows that the Oracle Gap is not absent, but merely hidden by aggregation.
\end{itemize}

\begin{takeawaybox}
    \textbf{Principle~\ding{202}.} Our findings suggest that benchmarks should be explicitly stratified by difficulty. Standard aggregate metrics are insufficient, as they allow high success rates on trivial tasks to mask significant failures on complex ones. Instead of relying on a single average score, reporting performance across distinct difficulty levels enables a more granular assessment. This approach is essential for exposing hidden generalization weaknesses and ensuring that reported improvements reflect robust reasoning rather than superficial fitting to easy data.
\end{takeawaybox}

\subsubsection{The Impact of Training Difficulty}
\label{sec:complexity-test}

\textbf{Setup and Phenomenon.}
To investigate the impact of training data difficulty on final generalization, we conduct a complexity test. We first train five generalist-optimized models, $M_{L_i}$ for $i \in \{1, \dots, 5\}$, on the previously defined difficulty-stratified training sets, $\mathcal{D}_{\text{train}}^{L_i}$. The key difference from our prior analysis lies in the evaluation protocol, which is centered around a novel, balanced test set.

\vspace{-1mm}
\begin{itemize}[leftmargin=*, topsep=3pt, itemsep=3pt]
    \item \textbf{\texttt{Test\_Balanced}:} 
    This is the unified and balanced evaluation suite, constructed by sampling an equal number of problems from each of the five difficulty levels. This results in a test set $\mathcal{D}_{bal}$ composed of five equal-sized partitions, $\{\mathcal{D}_{\text{test, bal}}^{L_j}\}_{j=1}^5$.
\end{itemize}
\vspace{-1mm}

Unlike the models in the first experiment, these models are "generalist-optimized," meaning we select the checkpoint for each $M_{L_i}$ with the highest overall accuracy on the \texttt{Test\_Balanced} set. We then analyze the relationship between the difficulty of the training data and the model's final average performance on this balanced benchmark.Detailed performance data for this experiment is provided in Appendix~\ref{app:Supplementary Experiment}.

\begin{figure}[t]
    \centering
    \includegraphics[width=\linewidth]{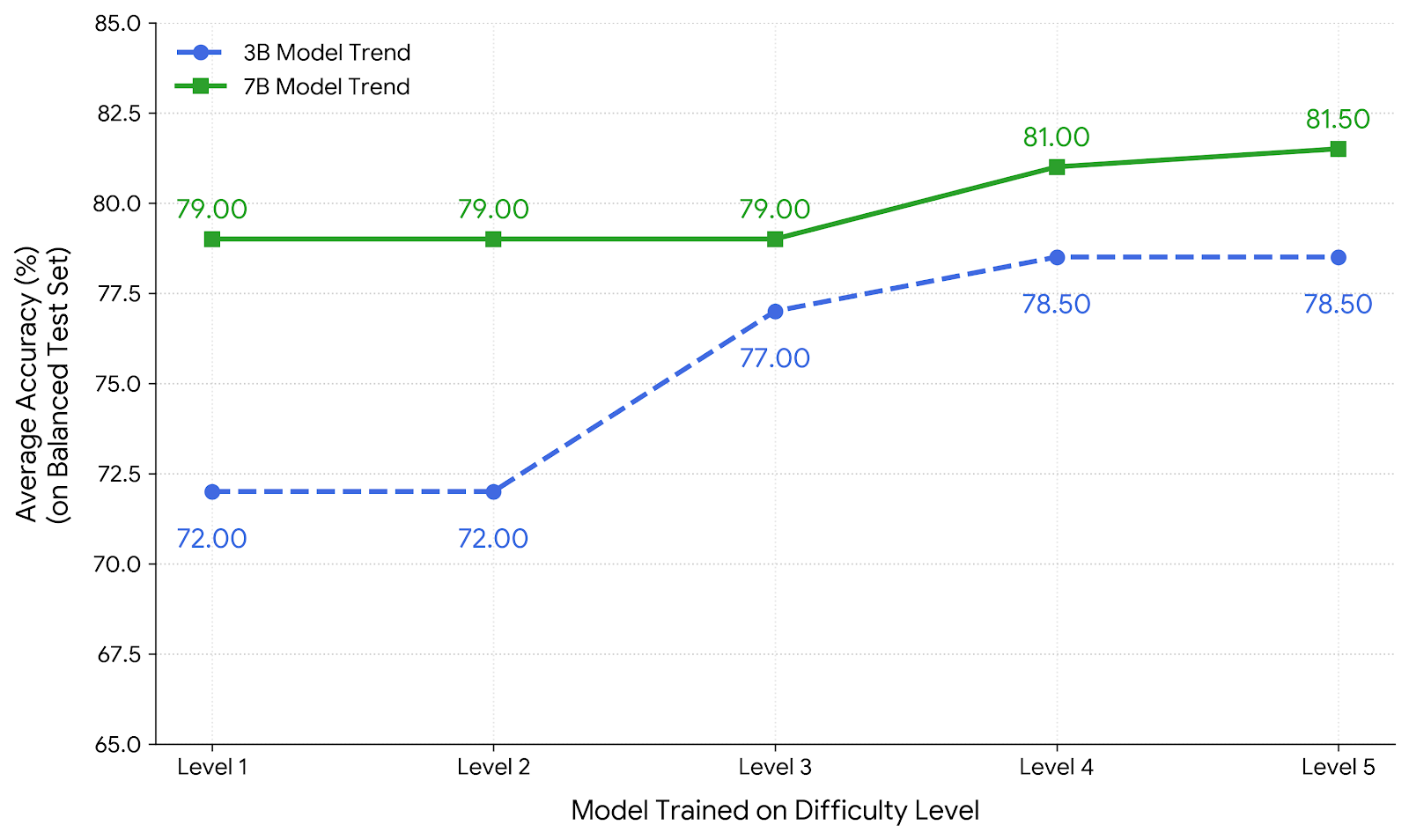}
    \caption{
        \textit{Average performance on the balanced test set.} 
        Consistent with Finding 3, we observe a positive correlation between training difficulty and generalization. Models trained on higher difficulty levels (L4--L5) consistently outperform those trained on easier data, yielding the strongest generalist performance.
    }
    \label{fig:cross-difficulty}
\end{figure}

\paragraph{Finding 3: Training on Difficult Problems Boosts Transferable Generalization.}
Our analysis reveals that models trained on higher difficulty levels (L4--L5) exhibit ''downward compatibility,'' effectively solving simple tasks while retaining complex reasoning capabilities. In contrast, models trained on easier data fail to generalize upward (Figure~\ref{fig:cross-difficulty}). Consequently, beyond evaluation considerations, these observations point to the potential benefit of including more challenging problems in training sets to encourage broader generalization.

\subsection{Evidence for Improved Benchmarking}
While the previous section highlighted the importance of problem complexity, difficulty is just one facet of generalization. To further probe our hypotheses regarding benchmark limitations and motivate our design principles, we introduce two stress tests for generalization forms: the Distribution Test (Section~\ref{sec:distribution-test}), quantifying brittleness to semantic shifts, and the Counterfactual Test (Section~\ref{sec:counterfactual-test}), distinguishing reasoning from memorization.
\subsubsection{The Distribution Test}
\label{sec:distribution-test}
\textbf{Setup.}
Standard train-test split typically evaluate models under an i.i.d. setting, in which generalization is assessed primarily within the training distribution. To evaluate generalization beyond this paradigm, we constructed a testbed using 44,785 problems from the benchmarks analyzed in Section~\ref{sec:Benchmark Flaws}. By clustering these via K-Means on embeddings, we modeled a spectrum from in-distribution to out-of-distribution scenarios based on semantic distance.

\begin{table*}[h]
    \centering
    \caption{\textit{Validation of Performance Inversion across Model Scales.} This table validates the brittleness of RL-tuned models using Global Cosine Distance. Both 3B and 7B models exhibit a clear "Performance Inversion" trend: while they show gains on semantically close data (d1), these gains diminish and eventually turn into significant penalties on distant, out-of-distribution data (d5).}
    \label{tab:distribution-test-formal-wide}

    \definecolor{gainGreen}{HTML}{008000}
    \definecolor{lossRed}{HTML}{C00000}

    \begin{tabular*}{\textwidth}{l @{\extracolsep{\fill}} cccccc}
        \toprule
        \textbf{Metric / Bin} & \textbf{d1 (Closest)} & \textbf{d2} & \textbf{d3} & \textbf{d4} & \textbf{d5 (Farthest)} & \textbf{Trend} \\
        \midrule
        
        \textbf{Qwen2.5-3B Gain} & 
        \textcolor{gainGreen}{+2.25\%} & 
        \textcolor{gainGreen}{+1.25\%} & 
        0.00\% & 
        \textcolor{lossRed}{-1.00\%} & 
        \textcolor{lossRed}{-3.75\%} & 
        $\searrow$ (Inverted) \\
        
        \textbf{Qwen2.5-7B Gain} & 
        \textcolor{gainGreen}{+7.25\%} & 
        \textcolor{gainGreen}{+6.50\%} & 
        \textcolor{gainGreen}{+5.00\%} & 
        \textcolor{gainGreen}{+1.25\%} & 
        \textcolor{lossRed}{-2.50\%} & 
        $\searrow$ (Inverted) \\
        
        \bottomrule
    \end{tabular*}
\end{table*}
\begin{itemize}[leftmargin=*, topsep=3pt, itemsep=3pt]

    \item \textbf{Core Training Set ($\mathcal{D}_{\text{core}}$):}
    We formed a concentrated training set by selecting the 2,000 problems closest to a cluster's centroid. This set was constructed primarily to serve as a distributional anchor to precisely identify OOD test sets based on semantic distance, while also simulating the strictly defined ``seen'' distribution of standard benchmarks(details in Appendix~\ref{app:distribution_test_data}).
    
    \item \textbf{Core-Trained Model ($M_{\text{core}}$):}
    We fine-tuned a specialist model exclusively on this dataset, formally $M_{\text{core}} \triangleq T(M_{base}, \text{RL}, \mathcal{D}_{\text{core}})$. This model is designed to be an expert solely on this specific distribution.
\end{itemize}

Finally, five test sets, $\{\mathcal{D}_{\text{test}}^{d_k}\}_{k=1}^5$, were constructed by sampling 80 problems each from the remaining data, which were binned according to increasing semantic distance $d_k$ from the $\mathcal{D}_{\text{core}}$ centroid.
\begin{Hypothesis}[title=Hypothesis 1: Optimization for Specific Distributions Induces Brittle Generalization]
We hypothesize that optimizing for a specific distribution yields brittle heuristics rather than robust skills. While we expect strong performance on data matching the training distribution (simulating benchmark conditions), we predict a \textbf{Performance Inversion} on out-of-distribution (OOD) data. We measure the specialist's gain over the baseline, $\text{Gain}(k) \triangleq P(M_{\text{core}}, \mathcal{D}_{\text{test}}^{d_k}) - P(M_{base}, \mathcal{D}_{\text{test}}^{d_k})$, testing if the advantage vanishes and eventually reverses as semantic distance $d_k$ increases:
\begin{equation}
    \exists k \in \{1, \dots, 5\} \quad \text{s.t.} \quad \text{Gain}(k) < 0
\end{equation}
Confirming this would demonstrate that high i.i.d. scores can mask harmful, non-generalizable biases instilled during fine-tuning.
\end{Hypothesis}

\paragraph{Finding 4: I.I.D. Test-Set Performance Is Not a Reliable Indicator of Generalization.}
Our distribution test (Table~\ref{tab:distribution-test-formal-wide}) confirms that excelling on a static distribution can be actively harmful to robustness. While the specialist model ($M_{\text{core}}$) dominates on in-distribution data (simulating high benchmark scores), this advantage is revealed to be brittle: it vanishes with semantic distance and culminates in a performance inversion on OOD sets. Here, the specialist’s accuracy collapses below that of the un-tuned baseline, demonstrating that what appears to be "capability" on i.i.d. test sets is often merely a harmful, non-generalizable bias.
\begin{takeawaybox}
    \textbf{Principle~\ding{203}.} Incorporating Distributional Robustness.
Our findings suggest that a faithful benchmark should go beyond in-distribution evaluation to actively probe for robustness
against distributional shifts. It should include a spectrum of out-of-distribution (OOD) challenges to penalize brittle, over-specialized models.
\end{takeawaybox}

\subsubsection{The Counterfactual Robustness Test}
\label{sec:counterfactual-test}

\textbf{Setup.}
To rigorously test whether our models perform genuine deductive reasoning or merely recite pre-trained knowledge, we designed a counterfactual robustness test. The experiment is constructed around the following key components:

\begin{itemize}[leftmargin=*, topsep=3pt, itemsep=3pt]

    \item \textbf{Test Sets ($\mathcal{D}_{bal}$ and $\mathcal{D}_{cf}$):}
Our experiment uses two test sets: a standard balanced set, $\mathcal{D}_{bal}$, from the MATH benchmark, and our primary evaluation set, $\mathcal{D}_{cf}$, which is created by transforming a subset of problems from $\mathcal{D}_{bal}$ (see Appendix~\ref{app:counterfactual_generation} for details).

    \item \textbf{Counterfactual Transformation ($c_{\text{real}} \to c_{\text{fake}}$):}
    The transformation process involves identifying a problem's core, real-world mathematical rule, $c_{\text{real}}$, and explicitly replacing it with a novel, contrary-to-fact premise, $c_{\text{fake}}$.
    
    \item \textbf{Evaluation Criterion:}
    A model's response is marked as correct only if it correctly and exclusively applies the explicitly stated counterfactual premise, $c_{\text{fake}}$. This strict criterion ensures we are measuring on-the-fly reasoning rather than answer correctness based on memorized knowledge.
    
\end{itemize}

We then evaluated our main RL-tuned models, \texttt{Qwen2.5-3B-MATH} and \texttt{Qwen2.5-7B-MATH}, on the new counterfactual set $\mathcal{D}_{cf}$.To assess the quality of the generated counterfactual set, we additionally conducted a human audit on 50 randomly selected samples, evaluated by three PhD students specializing in LLMs. The audit found that 93.34\% of samples were unambiguous and 94.67\% were solvable under the stated counterfactual rule (Appendix~\ref{app:counterfactual_audit}).
\begin{Hypothesis}[title=Hypothesis 2: Models Prioritize Recitation Over Reasoning]
We test the hypothesis that models default to reciting memorized knowledge ($c_{\text{real}}$) instead of reasoning from a novel premise ($c_{\text{fake}}$). A confirmation is indicated by a significant performance collapse on the counterfactual set, formalized as:
\begin{equation}
    P(M_{\text{RL}, train}, \mathcal{D}_{cf}) \ll P(M_{\text{RL}, train}, \mathcal{D}_{bal})
\end{equation}
\end{Hypothesis}

\begin{table}[t]
    \centering
    \caption{Performance collapse under standard and counterfactual evaluation.}
    \label{tab:counterfactual-results}
    
    \begin{tabular*}{\columnwidth}{l @{\extracolsep{\fill}} cc}
        \toprule
        \textbf{Model} & \makecell{$\mathcal{D}_{bal}$ \\ (\%)} & \makecell{$\mathcal{D}_{cf}$ \\ (\%)} \\
        \midrule
        3B-MATH & 64.20 & 36.00 \\
        7B-MATH & 74.80 & 41.20 \\
        \bottomrule
    \end{tabular*}
\end{table}

\paragraph{Finding 5: Counterfactual Failures Reveal Recitation Over Reasoning.}
Our counterfactual robustness test (Table~\ref{tab:counterfactual-results}) reveals a critical failure in models' ability to reason from novel premises. This is quantitatively evident in the severe performance degradation on the counterfactual test, where accuracies for our 7B and 3B models drop from 74.80\% and 64.20\% to 41.20\% and 36.00\%, respectively. A qualitative analysis of the model's chain-of-thought process confirms the cause of this failure (see Appendix~\ref{app:PESAMD} for a detailed example). When presented with a problem that redefines the order of operations to PESAMD, the model completely disregards the new rule and defaults to the standard PEMDAS operations it has memorized. This provides definitive evidence that it operates as a pattern-matching engine that recites knowledge, rather than as a flexible, deductive reasoner.

\vspace{-2mm}
\begin{takeawaybox}
    \textbf{Principle \ding{204}: } Assessing Counterfactual Reasoning.
    A faithful benchmark requires distinguishing true deduction from mere recitation. Our counterfactual test highlights a critical failure mode: when faced with novel, contrary-to-fact rules, models consistently default to reciting memorized knowledge rather than applying the new premise. Consequently, to penalize this brittle behavior and reward flexible reasoning, effective evaluation entails including problems that create a direct conflict between memorized priors and on-the-fly deduction.
\end{takeawaybox}

\section{Related work}
\label{sec:related_works}

\subsection{Reasoning in Large Language Models}
Chain-of-Thought (CoT) prompting has become a cornerstone for eliciting complex reasoning in Large Language Models (LLMs) \citep{wei_chain--thought_2022,zelikman2022star}. Along with advanced strategies like Tree of Thoughts~\citep{wang2022self, yao_tree_2023, yu2025frame}, these approaches improve reasoning by guiding models to generate step-by-step rationales. Furthermore, fine-tuning on high-quality reasoning datasets remains a critical method for instilling these skills directly into model parameters \citep{NEURIPS2022_18abbeef}. While these efforts have driven remarkable performance improvements on popular benchmarks, our work diverges from this trend of score optimization. Instead, we critically interrogate the benchmarks themselves, arguing that the resulting gains are often an illusion created by structural flaws rather than a sign of true reasoning acquisition.
\subsection{Reinforcement Learning for LLM}
To overcome the limitations of Supervised Fine-Tuning (SFT), Reinforcement Learning (RL) actively optimizes LLMs by directly rewarding correct outcomes \citep{10.5555/3600270.3602281}. Foundational algorithms like PPO~\citep{schulman2017proximal} and stability-enhancing methods like GRPO~\citep{shao_deepseekmath_2024}—which leverages group-based comparisons—have significantly boosted benchmark scores. Recent work has also explored reducing interference during alignment in multi-objective settings \citep{lin_orthalign_2025}. However, the reliance on outcome rewards is contested by process-based approaches that scrutinize reasoning steps to ensure true understanding \citep{li2025think}. Our work adds to this discourse by demonstrating that, even with advanced RL, structural benchmark flaws often lead to the reinforcement of brittle, non-generalizable behaviors.

\subsection{Analysis and Critique of Benchmarks}
While benchmarks like GSM8K \citep{cobbe2021gsm8k} and MATH \citep{hendrycks_measuring_2021} are vital for driving progress, research increasingly shows that models often exploit dataset artifacts and ``shortcuts'' rather than learning robust skills \citep{geirhos2020shortcut}. This has motivated rigorous evaluation methods, such as testing on out-of-distribution (OOD) or adversarially perturbed examples \citep{jia-liang-2017-adversarial}, to probe true generalization. However, these methods typically stress model capabilities without diagnosing the underlying benchmark properties that permit such brittle learning. We contribute to this critical analysis by introducing diagnostic tools—specifically the Oracle Performance Gap (OPG) and difficulty-stratified evaluations—that provide quantitative evidence revealing high scores as illusions of capability, thereby motivating our proposed principles.
\section{Conclusion}
\label{sec:conclusion}
In this work, we critically analyze RL-based reasoning benchmarks, arguing that high scores may often reflect brittle shortcut learning rather than robust generalization. Our empirical framework, anchored by the OPG diagnostic, reveals structural limitations such as data homogeneity and redundancy. Furthermore, stress tests suggest substantial model fragility, as evidenced by asymmetric generalization across difficulty levels and failures on counterfactual tasks. Finally, we distill our findings into three principles for next-generation benchmarks: \textit{difficulty stratification}, \textit{distributional robustness}, and \textit{counterfactual reasoning}.
\section*{Limitations}

\noindent \textbf{Model architectures} evaluated in our main experiments are still concentrated on the Qwen2.5 family (3B and 7B). Although we additionally validated the near-zero OPG trend on another open-weight model family (Llama-3-8B) and under an alternative RL algorithm (DAPO), we have not exhaustively tested a broader range of open-weight architectures or closed-source frontier models.

\noindent \textbf{Task domains} in our primary analysis remain centered on reasoning-intensive settings, especially mathematics and related benchmarks. Although we further extended the evaluation to additional reasoning domains such as HotpotQA, MedQA, and LogiQA, it remains unclear whether the same pattern persists in modalities with substantially different evaluation dynamics, such as code generation or creative writing. Future work is needed to determine whether our findings reflect a broader trend in RL fine-tuning or remain concentrated in reasoning-heavy tasks.

\noindent \textbf{Semantic partitioning} for our Distribution Test relies on specific tools---the \texttt{all-mpnet-base-v2} sentence encoder and K-Means clustering---to define data splits. While we employed rigorous metrics like Silhouette scores to validate these clusters, alternative embedding models or distance metrics could potentially yield different semantic boundaries.

\noindent \textbf{Automated data generation} for the Counterfactual Robustness Test relies on a pipeline driven by Gemini 2.5 Pro\citep{comanici_gemini_2025}. While this enabled large-scale evaluation, the diversity and complexity of the generated counterfactual rules remain bounded by the capabilities of the generator model. Although we supplemented this pipeline with a human audit of sample quality, broader validation of counterfactual diversity and quality remains an important direction for future work. As noted in our Ethics Statement, we also did not conduct a full audit of latent societal biases in the datasets or models used.

\section*{Ethical Considerations}

This work is motivated by the need to improve the scientific rigor of reinforcement learning evaluation. By examining the ``illusion of capability'' in current benchmarks, we aim to support the development of more robust and trustworthy reasoning systems.

\noindent \textbf{Broader Impact and Trustworthiness.} The deployment of RL-tuned models that perform well on static benchmarks but fail to generalize may pose risks in real-world applications, particularly in reasoning-intensive domains. Our findings highlight that high benchmark scores can mask brittle behaviors induced by over-specialization. By proposing stricter evaluation principles---such as distributional robustness and counterfactual testing---we advocate for evaluation settings that better distinguish robust reasoning from superficial pattern matching.

\noindent \textbf{Data Usage and Compliance.} Our experiments use publicly available academic datasets, including MATH, GSM8K, HeadQA, HotpotQA, MedQA, and LogiQA, together with a compiled in-house dataset (DeepScaler) derived from public sources. All data was used solely for academic research purposes. We acknowledge that these datasets may inadvertently contain sensitive information or reflect historical biases. While we did not conduct a full audit of latent societal biases in the external datasets used in this study, we recognize this as an important direction for future work on equitable model evaluation.

\noindent \textbf{Computational Resources.} The experiments were conducted on a single server with 4 NVIDIA A100 GPUs. We adhered to efficient training practices to minimize unnecessary computational costs and environmental impact.

\noindent \textbf{AI Assistance Declaration.} In accordance with conference policies, we state that Large Language Models (specifically Gemini 2.5 Pro) were used in this work. Their use was limited to two functions: (1) supporting automated data annotation and counterfactual generation within our experimental pipeline, and (2) assisting with grammatical refinement and language polishing of the manuscript. All scientific concepts, experimental designs, and core intellectual contributions originated from the human authors.

\section*{Acknowledgements}
We sincerely thank all the anonymous reviewers and (S)ACs for their constructive comments and helpful suggestions. This work was supported by the National Key Research and Development Program of China (Grant No.~2024YFF0507603) and the Anhui Provincial Major Science and Technology Project (Nos.~202303a07020006 and 202304a05020071).

\bibliography{custom}
\bibliographystyle{acl_natbib}

\clearpage
\appendix
\section{Full Evaluation Setup}
\label{sec:exp_setup}
\subsection{Post-training Methods}
\textbf{Reinforcement Learning}
Reinforcement Learning (RL) has recently proven effective at steering large language models toward complex, multi-step objectives by optimizing policies with scalar reward signals~\citep{zeng2025simplerl}. For our main experiments, we use the \texttt{easy-r1} framework, a fork of the original \texttt{veRL} project~\citep{zheng2025easyr1}. We adopt its implementation of Group Relative Policy Optimization (GRPO)~\citep{shao_deepseekmath_2024} to fine-tune \texttt{Qwen2.5-7B-Instruct}, using final-answer correctness as the reward signal. Our RL configuration uses a learning rate of $1 \times 10^{-6}$ with the AdamW optimizer and a weight decay of $1.0 \times 10^{-2}$. We generate 5 responses per prompt with a maximum sequence length of 4096 tokens, using a temperature of 1.0 and a top-$p$ of 0.99. The model is updated with a global batch size of 16. KL-divergence regularization is enabled with a coefficient of $1.0 \times 10^{-2}$. We train for 5 epochs and select the checkpoint with the best validation performance. Additional robustness experiments involving alternative algorithms, architectures, and inference settings are described in Appendix~\ref{app:opg_robustness}.

\noindent \textbf{Supervised Fine-Tuning}
Supervised Fine-Tuning (SFT) remains a fundamental technique for adapting large pre-trained models by directly minimizing cross-entropy on high-quality datasets~\citep{parashar_curriculum_2026}. We use the LLaMA-Factory framework \citep{zheng2024llamafactory}, which is an extensible and user-friendly framework supporting multiple architectures and advanced optimization algorithms, to fine-tune our model on teacher-generated chain-of-thought traces. We use $1 \times 10^{-6}$  as learning rate, the batch size is 512 and we train for 5 epoch to align with our RL settings.

\subsection{Datasets and Benchmarks}
Our primary analysis was conducted on the following four benchmarks, chosen to cover a range of mathematical and general reasoning tasks. Additional reasoning domains used in our robustness analyses are described in Appendix~\ref{app:opg_robustness}.

\begin{itemize}[leftmargin=*, topsep=3pt, itemsep=2pt]
    \item \textbf{MATH}~\citep{hendrycks_measuring_2021}: A challenging dataset of 12,500 competition mathematics problems designed to test mathematical problem solving.
    \item \textbf{GSM8K}~\citep{cobbe2021gsm8k}: A dataset of 8,500 high-quality, linguistically diverse grade-school math word problems created to measure multi-step reasoning.
    \item \textbf{HeadQA}~\citep{vilares2019head}: A multiple-choice question answering dataset sourced from Spanish medical board exams, covering a wide range of topics and requiring specialized knowledge.
    \item \textbf{DeepScaler}: A proprietary in-house dataset created to evaluate specific mathematical reasoning abilities. It contains approximately 40,000 unique math problem-answer pairs compiled from sources such as AIME, AMC, Omni-MATH, and Still.
\end{itemize}

\subsection{Implementation Details}
All experiments were conducted on a single server equipped with 4 NVIDIA A100 (80GB) GPUs. Our implementation relies on PyTorch and the Hugging Face Transformers library.

\section{Detailed Data for Difficulty-Stratified Analysis}
\label{app:difficulty_data}
\subsection{Automated Difficulty Level Annotation}
\label{app:difficulty_annotation}

To ensure a systematic and reproducible partitioning of our datasets into difficulty levels (L1-L5), we employed an automated annotation pipeline. Instead of relying on subjective manual labeling, we developed a detailed rubric based on the cognitive complexity required for each problem and used a large language model (\texttt{Gemini 2.5 Pro}) to assign a difficulty score to each problem in our corpus.

The process was guided by the five-level standard defined below. For each problem, the full text of this rubric was provided to the LLM, which was then prompted to return the single most appropriate difficulty level.
\begin{table*}[t]
    \centering
    
    \caption{Cross-Difficulty Generalization Performance Matrix for the \textit{Qwen2.5-3B-Instruct} model. All values are pass@1 accuracy.}
    \label{tab:difficulty-matrix-3b-appendix}
    \begin{tabular*}{\textwidth}{l @{\extracolsep{\fill}} ccccc c}
        \toprule
        \multirow{2}{*}{\makecell[l]{\textbf{Trained on}}} & \multicolumn{6}{c}{\textbf{Evaluated on Training Set of Level}} \\
        \cmidrule(lr){2-7}
        & \textbf{Level 1} & \textbf{Level 2} & \textbf{Level 3} & \textbf{Level 4} & \textbf{Level 5} & \textbf{Average} \\
        \midrule
        \textbf{Level 1}   & 94.50\% & 85.00\% & 71.00\% & 66.00\% & 41.00\% & 71.50\% \\
        \textbf{Level 2}   & 93.00\% & 87.50\% & 73.00\% & 65.00\% & 42.50\% & 72.20\% \\
        \textbf{Level 3}   & 92.50\% & 86.00\% & 75.00\% & 66.00\% & 40.00\% & 71.90\% \\
        \textbf{Level 4}   & 92.50\% & 86.50\% & 72.00\% & 68.00\% & 43.00\% & 72.40\% \\
        \textbf{Level 5}   & 94.00\% & 87.00\% & 73.00\% & 62.00\% & 46.50\% & 72.50\% \\
        \midrule
        \textbf{Original}  & 92.00\% & 83.50\% & 69.50\% & 62.50\% & 43.50\% & 70.20\% \\
        \bottomrule
    \end{tabular*}

    \vspace{1em}

    \caption{Cross-Difficulty Generalization Performance Matrix for the \textit{Qwen2.5-7B-Instruct} model. All values are pass@1 accuracy.}
    \label{tab:difficulty-matrix-7b-appendix}
    \begin{tabular*}{\textwidth}{l @{\extracolsep{\fill}} ccccc c}
        \toprule
        \multirow{2}{*}{\makecell[l]{\textbf{Trained on}}} & \multicolumn{6}{c}{\textbf{Evaluated on Training Set of Level}} \\
        \cmidrule(lr){2-7}
        & \textbf{Level 1} & \textbf{Level 2} & \textbf{Level 3} & \textbf{Level 4} & \textbf{Level 5} & \textbf{Average} \\
        \midrule
        \textbf{Level 1}   & 97.00\% & 90.00\% & 78.00\% & 76.00\% & 52.00\% & 78.60\% \\
        \textbf{Level 2}   & 94.00\% & 91.50\% & 82.50\% & 76.00\% & 54.00\% & 79.60\% \\
        \textbf{Level 3}   & 95.50\% & 91.00\% & 83.50\% & 72.50\% & 56.50\% & 79.80\% \\
        \textbf{Level 4}   & 93.50\% & 88.50\% & 81.00\% & 80.00\% & 57.00\% & 80.00\% \\
        \textbf{Level 5}   & 94.50\% & 91.00\% & 78.00\% & 73.00\% & 64.00\% & 80.10\% \\
        \midrule
        \textbf{Original}  & 95.50\% & 87.50\% & 76.50\% & 74.00\% & 52.00\% & 77.60\% \\
        \bottomrule
    \end{tabular*}
\end{table*}
\begin{description}[leftmargin=*, style=unboxed, topsep=3pt, itemsep=3pt]

    \item[Level 1: Direct Application of Basic Rules.]
    Problems that can be solved in one or two steps, where each step is a direct application of a basic formula or operational rule. The solution path is linear and requires minimal strategic planning.

    \item[Level 2: Identification of Standard Models.]
    Problems that require identifying the correct standard model or general formula from a set of known methods. This tests for "pattern recognition" of classic problem types.

    \item[Level 3: Multi-Step, Cross-Conceptual Planning.]
    Problems that cannot be solved by a single standard model and require a coherent plan that links multiple concepts or steps, often from different mathematical areas.

    \item[Level 4: Application of Abstract Concepts.]
    Problems requiring a deep understanding and flexible application of a major, abstract mathematical theory. The solution process is often non-intuitive and relies on a foundational result within a branch of mathematics.

    \item[Level 5: Axiomatic Reasoning and Creation.]
    Problems that require reasoning "from first principles" within an axiomatic framework. This involves performing logical deductions, constructing proofs, or finding counterexamples based on the foundational rules of a mathematical structure.

\end{description}

The entire dataset was processed using a parallelized script with a thread pool executor to efficiently query the LLM API. The script included robust error handling and checkpointing to ensure the complete and accurate annotation of the corpus.
\subsection{Result}
\begin{table*}[t]
    \centering
    
    \caption{Performance of \textit{Qwen2.5-7B} generalist-optimized models on the balanced test set. Each row represents a model trained on a specific difficulty level ($L_i$), evaluated across test questions of all five difficulty levels.}
    \label{tab:complexity-test-7b-appendix}
    \begin{tabular*}{\textwidth}{l @{\extracolsep{\fill}} ccccc c}
        \toprule
        \multirow{2}{*}{\makecell[l]{\textbf{Trained on}}} & \multicolumn{6}{c}{\textbf{Evaluated on Test Set Questions of Level}} \\
        \cmidrule(lr){2-7}
        & \textbf{Level 1} & \textbf{Level 2} & \textbf{Level 3} & \textbf{Level 4} & \textbf{Level 5} & \textbf{Average} \\
        \midrule
        \textbf{Level 1}   & 97.50\% & 90.00\% & 82.50\% & 75.00\% & 50.00\% & 79.00\% \\
        \textbf{Level 2}   & 95.00\% & 90.00\% & 80.00\% & 77.50\% & 47.50\% & 79.00\% \\
        \textbf{Level 3}   & 97.50\% & 85.00\% & 85.00\% & 77.50\% & 50.00\% & 79.00\% \\
        \textbf{Level 4}   & 97.50\% & 87.50\% & 85.00\% & 80.00\% & 55.00\% & 81.00\% \\
        \textbf{Level 5}   & 97.50\% & 92.50\% & 82.50\% & 82.50\% & 52.50\% & 81.50\% \\
        \midrule
        \textbf{Original}  & 97.50\% & 87.50\% & 82.50\% & 77.50\% & 50.00\% & 79.00\% \\
        \bottomrule
    \end{tabular*}

    \vspace{1em}

    \caption{Performance of \textit{Qwen2.5-3B} generalist-optimized models on the balanced test set. The performance decay for models trained on easy levels (L1, L2) is particularly pronounced.}
    \label{tab:complexity-test-3b-appendix}
    \begin{tabular*}{\textwidth}{l @{\extracolsep{\fill}} ccccc c}
        \toprule
        \multirow{2}{*}{\makecell[l]{\textbf{Trained on}}} & \multicolumn{6}{c}{\textbf{Evaluated on Test Set Questions of Level}} \\
        \cmidrule(lr){2-7}
        & \textbf{Level 1} & \textbf{Level 2} & \textbf{Level 3} & \textbf{Level 4} & \textbf{Level 5} & \textbf{Average} \\
        \midrule
        \textbf{Level 1}   & 97.50\% & 82.50\% & 75.00\% & 72.50\% & 32.50\% & 72.00\% \\
        \textbf{Level 2}   & 95.00\% & 87.50\% & 80.00\% & 65.00\% & 35.00\% & 72.00\% \\
        \textbf{Level 3}   & 97.50\% & 90.00\% & 80.00\% & 72.50\% & 45.00\% & 77.00\% \\
        \textbf{Level 4}   & 95.00\% & 87.50\% & 87.50\% & 75.00\% & 47.50\% & 78.50\% \\
        \textbf{Level 5}   & 95.00\% & 87.50\% & 87.50\% & 75.00\% & 47.50\% & 78.50\% \\
        \midrule
        \textbf{Original}  & 92.50\% & 87.50\% & 77.50\% & 65.00\% & 22.50\% & 69.00\% \\
        \bottomrule
    \end{tabular*}
\end{table*}
\label{app:c_result}
This section provides the full cross-difficulty generalization performance matrices that form the basis for the analysis in Section~\ref{sec:average_score} and the visualizations in Figure~\ref{fig:final_comparison}. Table~\ref{tab:difficulty-matrix-3b-appendix} and Table~\ref{tab:difficulty-matrix-7b-appendix} present the results for the \texttt{Qwen2.5-3B-Instruct} and \texttt{Qwen2.5-7B-Instruct} models, respectively.

The data highlights two key phenomena discussed in the main text. First, asymmetric generalization is evident: in Table~\ref{tab:difficulty-matrix-7b-appendix}, the model trained on Level 5 achieves 94.50\% on Level 1, while the model trained on Level 1 only achieves 52.00\% on Level 5. This pattern is mirrored in the 3B model, confirming that training on complex tasks induces a ``downward compatibility'' that simple training lacks. Second, the average score proves deceptive. As shown in the `Average` column, scores across specialists are remarkably similar (e.g., 78.60\%--80.10\% for 7B). This similarity masks a critical distinction: low-difficulty specialists inflate their averages via simple tasks, whereas high-difficulty specialists achieve their scores through genuine robustness, a nuance the aggregate metric fails to capture.

\section{A Supplementary Experiment to the Difficulty Test}
\begin{figure*}[h!]
    \centering
    \begin{subfigure}[b]{0.48\textwidth}
        \centering
        \includegraphics[width=\textwidth]{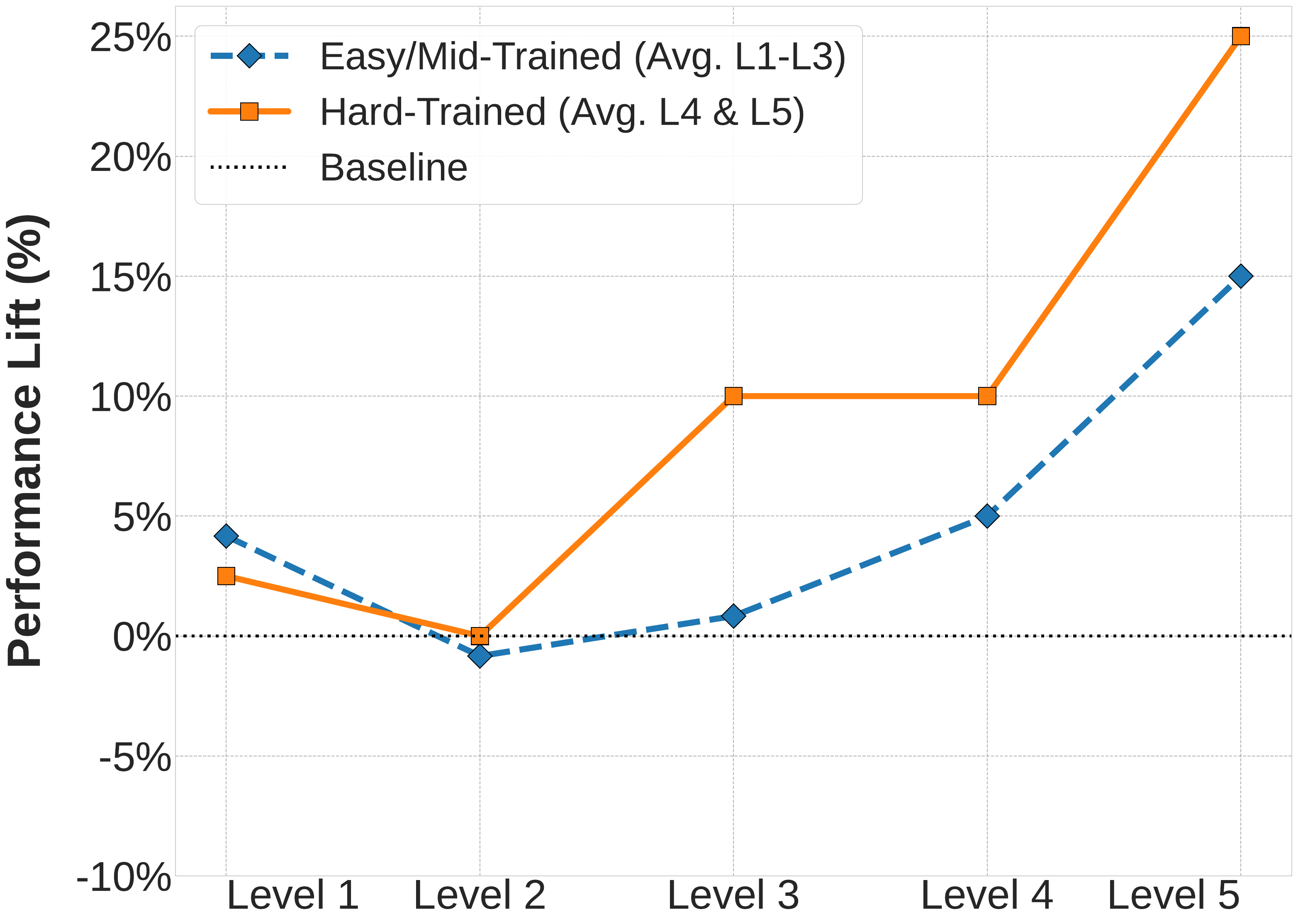}
        \caption{Performance lift of the 3B model.}
    \end{subfigure}
    \hfill 
    \begin{subfigure}[b]{0.48\textwidth}
        \centering
        \includegraphics[width=\textwidth]{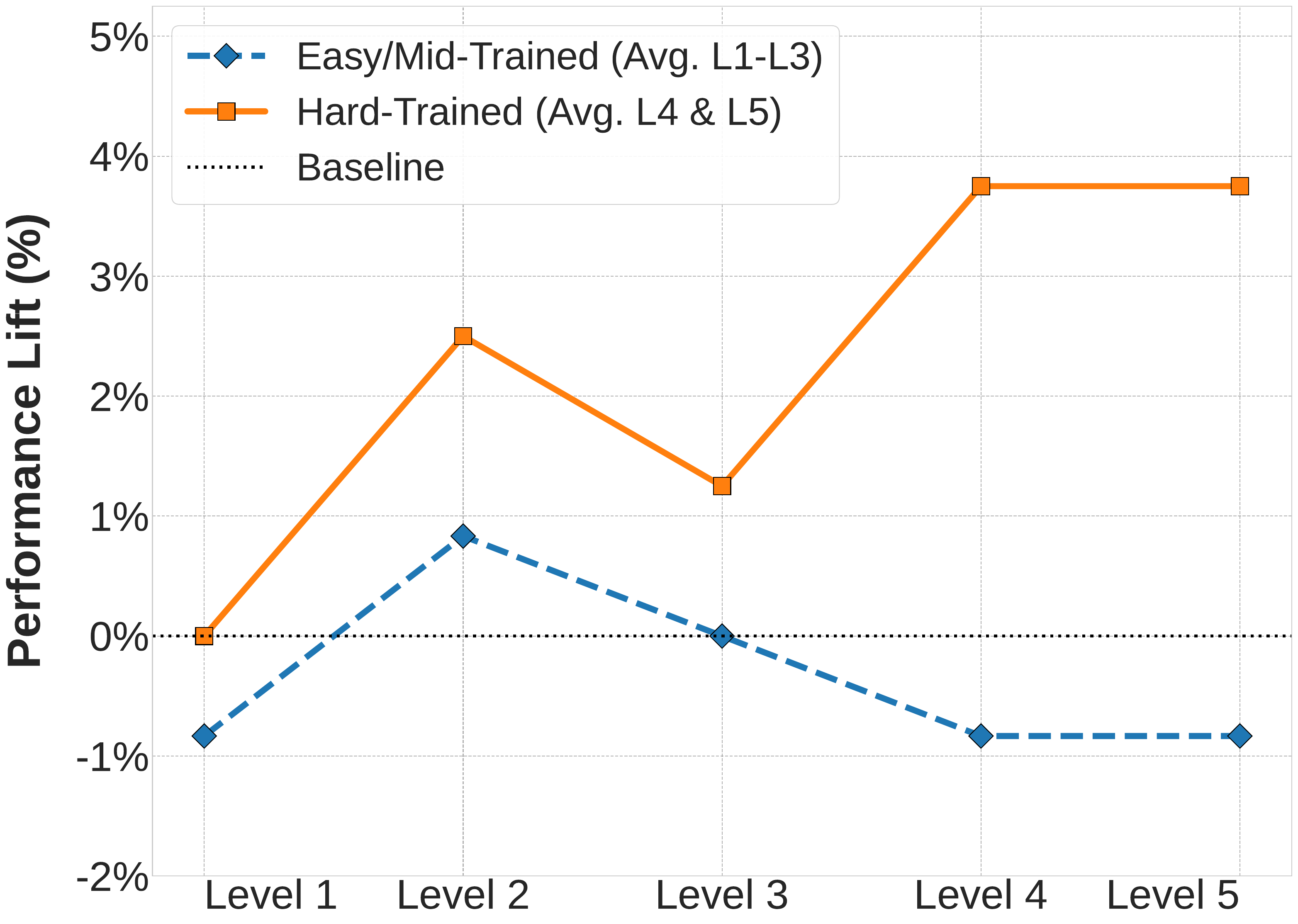}
        \caption{Performance lift of the 7B model.}
    \end{subfigure}
    \caption{
    \textit{Asymmetric Generalization is consistent across model scales. }
    Across both the 3B model (a) and the 7B model (b), training on high-difficulty problems (L4-L5, orange line) yields a uniformly superior performance lift over training on easier problems (L1-L3, blue line), proving that mastering complexity is essential for acquiring robust, transferable skills.Full performance data is provided in Table~\ref{tab:complexity-test-3b-appendix} and Table~\ref{tab:complexity-test-7b-appendix}.
    }
    \label{fig:complexity_lift_curves} 
\end{figure*}

This appendix provides the full performance data for the "generalist-optimized" models described in our supplementary experiment on the difficulty test. The performance lift curves presented in Figure~\ref{fig:complexity_lift_curves} in the main text are directly derived from the raw accuracy scores presented here. Table~\ref{tab:complexity-test-7b-appendix} details the results for the 7B model, while Table~\ref{tab:complexity-test-3b-appendix} shows the results for the 3B model.

\label{app:Supplementary Experiment}
\textbf{Setup.}
To investigate the impact of training data difficulty on final generalization, we conduct a complexity test. We first train five generalist-optimized models, $M_{L_i}$ for $i \in \{1, \dots, 5\}$, on the previously defined difficulty-stratified training sets, $\mathcal{D}_{\text{train}}^{L_i}$. The key difference from our prior analysis lies in the evaluation protocol, which is centered around a novel, balanced test set.

\vspace{-1mm}
\begin{itemize}[leftmargin=*, topsep=3pt, itemsep=3pt]
    \item \textbf{\texttt{Test\_Balanced}:} 
    This is the unified and balanced evaluation suite, constructed by sampling an equal number of problems from each of the five difficulty levels. This results in a test set $\mathcal{D}_{bal}$ composed of five equal-sized partitions, $\{\mathcal{D}_{\text{test, bal}}^{L_j}\}_{j=1}^5$.
\end{itemize}
\vspace{-1mm}
Unlike the models in the first experiment, these models are "generalist-optimized," meaning we select the checkpoint for each $M_{L_i}$ with the highest overall accuracy on the \texttt{Test\_Balanced} set.

Our complexity test reveals a stark pattern of asymmetric generalization, as illustrated in Figure~\ref{fig:complexity_lift_curves}. Models trained on high-difficulty problems (L4-L5) demonstrate a uniformly superior performance profile, outperforming their counterparts trained on easier data (L1-L3) across all evaluated task complexities. This suggests that complex reasoning skills naturally encompass the logic required for simpler tasks, whereas the reverse is not true. This finding has a critical implication for how we create datasets to train capable models: the training data must include a significant proportion of difficult problems. Therefore, for benchmark suites to drive meaningful progress, it is crucial that their provided training sets are sufficiently challenging to promote the development of truly robust models rather than merely encouraging the fitting of simple patterns.The data in these tables clearly illustrates this "asymmetric generalization" phenomenon. For example, in Table~\ref{tab:complexity-test-3b-appendix}, the model trained on Level 1 ($M_{L_1}$) achieves high accuracy (97.50\%) on Level 1 test problems but sees its performance drop to just 32.50\% on Level 5 problems, indicating that the model relies on shallow heuristics that collapse under increased cognitive load. In contrast, the model trained on Level 5 ($M_{L_5}$) maintains robust performance across all levels, demonstrating a more generalizable capability that effectively transfers downwards to easier tasks.

\section{Data Construction Protocol for the Distribution Test}
\label{app:distribution_test_data}

This section details the step-by-step procedure used to construct the specialized training and test sets for the Distribution Test, as described in Section~\ref{sec:distribution-test}. The entire process is designed to create a controlled environment for measuring generalization as a function of semantic distance. The process consists of three main stages:
\vspace{-3mm}
\paragraph{Step 1: Semantic Embedding and Clustering.}
We began with our full corpus of approximately 44,785 mathematics problems. To understand their semantic relationships, we first encoded each problem into a high-dimensional vector representation using the \texttt{all-mpnet-base-v2} sentence encoder. We then applied K-Means clustering to this high-dimensional embedding space. Using a combination of the Elbow method and Silhouette score analysis, we determined the optimal number of clusters to be $k=3$, effectively partitioning the entire dataset into three broad, semantically coherent groups.
\vspace{-3mm}
\paragraph{Step 2: Core Training Set (\texttt{Train\_Core}) Selection.}
Our goal was to create a highly concentrated, semantically narrow training set. To achieve precise semantic selection, we performed the analysis using Global Cosine Distance directly on the original high-dimensional embeddings, avoiding potential information loss from dimensionality reduction. We first calculated the centroid of a single target cluster (e.g., Cluster 1) within the high-dimensional space. We then computed the cosine distance between this centroid and all data points in the cluster. The 2,000 problems with the smallest cosine distances—representing the points semantically closest to the cluster center—were selected to form our exclusive \texttt{Train\_Core} training set.

\vspace{-3mm}
\paragraph{Step 3: Distance-Stratified Test Set Construction.}
To systematically construct test sets representing a gradient of increasing semantic distance, we leveraged the remaining pool of 42,785 problems explicitly excluded from \texttt{Train\_Core}. We first defined a stable reference origin by calculating the geometric centroid of the 2,000 \texttt{Train\_Core} vectors within the high-dimensional embedding space. Subsequently, for every candidate point in the hold-out corpus, we computed its cosine distance relative to this core centroid to quantify its semantic divergence. All candidate points were then sorted by distance and stratified into five equal-sized bins (quintiles). Finally, to ensure a balanced evaluation, we randomly sampled 80 distinct problems from each bin to create our five final test sets, D1 (semantically closest) through D5 (semantically farthest).

\begin{figure*}[h]
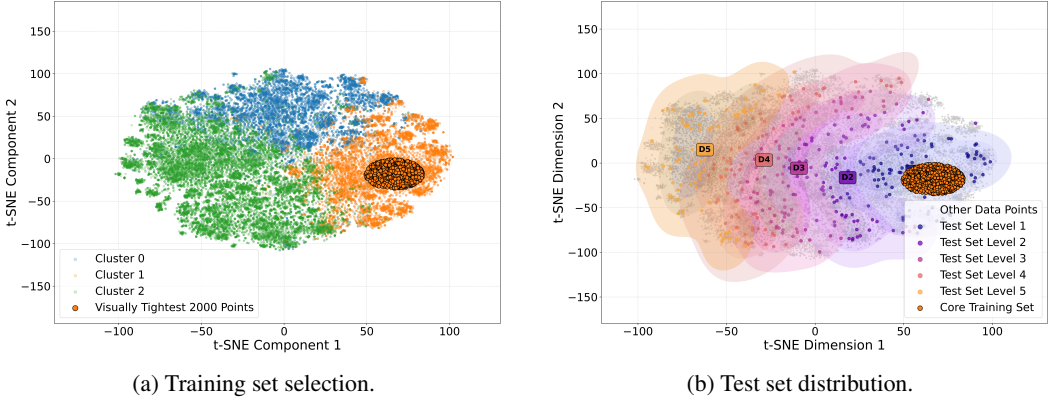
 
    \centering
    
    \begin{subfigure}[b]{0.48\textwidth}
        \centering
        \includegraphics[width=\textwidth]{visual_tightest_visualization_cluster_1.pdf}
        \caption{Training set selection.}
        \label{fig:train-set-viz}
    \end{subfigure}
    \hfill 
    \begin{subfigure}[b]{0.48\textwidth}
        \centering
        \includegraphics[width=\textwidth]{final_visualization_with_kde.pdf}
        \caption{Test set distribution.}
        \label{fig:test-set-viz}
    \end{subfigure}
    \caption{\textit{Visualization of the experimental data construction for the distribution test.} (a) The highly concentrated $\mathcal{D}_{\text{core}}$ set is selected from a semantic cluster. (b) The test sets are sampled and binned based on their increasing semantic distance from the $\mathcal{D}_{\text{core}}$ centroid.}
    \label{fig:data-selection}
\end{figure*}

The entire data construction pipeline is visually summarized in Figure~\ref{fig:data-selection}. Panel (a) illustrates the outcome of the \texttt{Train\_Core} selection process described in Step 2, while Panel (b) shows the resulting distribution of the five distance-stratified test sets as detailed in Step 3.

\section{ The Counterfactual Robustness Test}
\label{app:counterfactual_cases}

This section provides detailed, qualitative examples of how fine-tuned models fail on counterfactual reasoning tasks, as discussed in Section~\ref{sec:counterfactual-test}. Each table analyzes a specific failure case, comparing the required reasoning path (based on the novel, counterfactual premise) with the model's actual thought process. These examples concretely illustrate the models' strong tendency to disregard explicit instructions and default to their pre-trained, memorized knowledge.
\subsection{Methodology: Automated Dataset Generation}
\label{app:counterfactual_generation}

To ensure the diversity and systematic nature of our counterfactual examples, we developed and executed the following automated pipeline, moving beyond manual creation.
\vspace{-3mm}
\paragraph{Step 1: Strategy — LLM as Data Creator.}
Our core strategy was to leverage a powerful Large Language Model to act as a creative research assistant. This approach allows for the large-scale and consistent application of complex transformation rules needed to create a high-quality counterfactual dataset.
\vspace{-3mm}
\paragraph{Step 2:Task Definition — The Counterfactual Transformation.}
We provided the LLM (\texttt{Gemini 2.5 Pro}) with a detailed, multi-step prompt that precisely defined the transformation task. The instructions guided the model to first analyze a given standard problem to identify a core logical or mathematical rule. Subsequently, the model was tasked to invent a plausible but contrary-to-fact rule, rewrite the problem statement to include this new premise, and finally, generate a new step-by-step solution based exclusively on the novel rule.
\vspace{-3mm}
\paragraph{Step 3: Execution — Parallelized Pipeline.}
This generation process was applied to our entire source dataset. To manage the scale, the pipeline was executed in parallel using a Python script with a \texttt{ThreadPoolExecutor} to handle concurrent API requests. The full, unabridged master prompt used in this process is available in our supplementary materials to ensure full reproducibility.

\subsection{Human Audit of Counterfactual Data Quality}
\label{app:counterfactual_audit}

To quantitatively assess the quality of the generated counterfactual dataset, we conducted a human audit on 50 randomly selected samples. The audit was performed by three PhD students specializing in large language models. Each sample was evaluated according to the following criteria:

\begin{itemize}[leftmargin=*, noitemsep, topsep=2pt]
    \item \textbf{Unambiguous:} The counterfactual rule is explicitly stated and clearly overrides the model's pre-trained prior or default rule.
    \item \textbf{Solvable:} The problem contains sufficient information for a logically valid solution under the stated counterfactual rule.
\end{itemize}

The audit found that 93.34\% of the sampled instances were judged to be unambiguous, and 94.67\% were judged to be solvable. These results provide supporting evidence that the generated counterfactual test set is of sufficiently high quality for evaluating whether models follow explicitly stated novel rules rather than defaulting to memorized priors.
\subsection{Case Study: Arithmetic Order of Operations}
\label{app:PESAMD}
\begin{premisebox}
A novel order of operations, \textbf{PESAMD}, is defined: Parentheses, Exponents, \textbf{S/A}, then \textbf{M/D}. The model is asked to evaluate $f(x) = \frac{3x-2}{x-2}$.
\end{premisebox}

\begin{comparisonbox}
    
    \textbf{Correct Reasoning (PESAMD)}
    \begin{enumerate}[nosep, leftmargin=*]
        \item \textbf{Numerator (S first):} $3 \times (0 - 2) = -6$
        \item \textbf{Denominator:} $0 - 2 = -2$
        \item \textbf{Division (last):} $\frac{-6}{-2} = 3$
    \end{enumerate}
    The final correct answer is $\mathbf{3}$.

    \tcblower

    \textbf{Model's Actual Reasoning}
    \begin{enumerate}[nosep, leftmargin=*]
    \item \textbf{Numerator (M first):}  $3 \times 0 = 0$, then $0-2 = -2$.
    \item \textbf{Denominator:} $0-2 = -2$.
    \item \textbf{Result:} $\frac{-2}{-2} = 1$.
\end{enumerate}
    The final incorrect answer is $\mathbf{1}$.
\end{comparisonbox}

\subsection{Case Study: Number Theory Divisor Rule}

\begin{premisebox}
A new system defines the number of divisors of $N=p_1^{a_1} \cdots$ as the \textbf{sum} of $(a_i+1)$ values. Find the number of divisors for $N=12$.
\end{premisebox}

\begin{comparisonbox}
    
    \textbf{Correct Reasoning (Sum Rule)}
    \begin{enumerate}[nosep, leftmargin=*]
        \item Prime factorization of 12 is $2^2 \times 3^1$.
        \item The exponents are $a_1=2, a_2=1$.
        \item Apply the new \textbf{sum rule}: $(2+1) + (1+1) = 5$.
    \end{enumerate}
    The final correct answer is $\mathbf{5}$.
    
    \tcblower

    \textbf{Model's Actual Reasoning}
    \begin{enumerate}[nosep, leftmargin=*]
        \item Correctly finds prime factorization: $12=2^2 \times 3^1$.
        \item {\color{red!80!black} Ignores the "sum" rule and applies the memorized "product" rule}: $(2+1) \times (1+1) = 6$.
    \end{enumerate}
    The final incorrect answer is $\mathbf{6}$.
\end{comparisonbox}

\subsection{Case Study: Physics Speed Formula}

\begin{premisebox}
A car travels 120 km in 2 hours. In this reality, 'average speed' is calculated as: \textbf{speed = time / distance}. Find the speed.
\end{premisebox}

\begin{comparisonbox}
    
    \textbf{Correct Reasoning (New Formula)}
    \begin{enumerate}[nosep, leftmargin=*]
        \item Identify Time = 2 hours, Distance = 120 km.
        \item Apply the new formula time / distance: $2 \div 120 = \frac{1}{60}$.
    \end{enumerate}
    The final correct answer is $\mathbf{\frac{1}{60}}$ \textbf{km/h}.
    
    \tcblower

    \textbf{Model's Actual Reasoning}
    \begin{enumerate}[nosep, leftmargin=*]
        \item Correctly identifies Time and Distance.
        \item {\color{red!80!black} Ignores the new formula and applies the memorized, standard formula} `distance / time`: $120 \div 2 = 60$.
    \end{enumerate}
    The final incorrect answer is $\mathbf{60}$ \textbf{km/h}.
\end{comparisonbox}
\section{Additional Robustness Analyses for OPG}
\label{app:opg_robustness}

This section reports additional robustness analyses for the Oracle Performance Gap (OPG). We test whether the near-zero OPG trend persists across different RL algorithms, model families, task domains, and inference settings. Across all these settings, OPG remains small.

\subsection{Across RL Algorithms: DAPO}
\label{app:dapo_evaluation}

We additionally evaluate DAPO to test whether the near-zero OPG trend extends beyond GRPO. Table~\ref{tab:dapo_results} shows that the gap between Standard RL and Oracle RL remains minimal across GSM8K, DeepScaler, and HeadQA.

\begin{table}[h]
    \captionsetup{justification=raggedright,singlelinecheck=false}
    \centering
    \caption{Near-zero OPG under an alternative RL algorithm (DAPO).}
    \label{tab:dapo_results}
    \vspace{2mm}
    \small
    \begin{tabularx}{\linewidth}{X c c c}
        \toprule
        \textbf{Benchmark} & \textbf{Train} & \textbf{Oracle} & \textbf{Gap} \\
        \midrule
        GSM8K      & 86.40\% & 86.90\% & 0.50\% \\
        DeepScaler & 43.85\% & 44.20\% & 0.35\% \\
        HeadQA     & 67.10\% & 67.78\% & 0.68\% \\
        \bottomrule
    \end{tabularx}
\end{table}

\subsection{Across Architectures and Domains}
\label{app:arch_domain_robustness}

We further test whether the near-zero OPG phenomenon extends beyond the primary Qwen2.5-based mathematical setting. Table~\ref{tab:llama_opg} shows similarly small OPG values on an additional open-weight model family, Llama-3-8B. Table~\ref{tab:domain_opg} shows that the same trend also holds on non-mathematical reasoning domains, including HotpotQA, MedQA, and LogiQA.

\begin{table}[h]
    \captionsetup{justification=raggedright,singlelinecheck=false}
    \centering
    \caption{OPG on Llama-3-8B.}
    \label{tab:llama_opg}
    \vspace{2mm}
    \small
    \begin{tabularx}{\linewidth}{X c c c}
        \toprule
        \textbf{Benchmark} & \textbf{Train} & \textbf{Oracle} & \textbf{OPG (\%)} \\
        \midrule
        GSM8K      & 87.45 & 88.40 & 1.07 \\
        MATH       & 43.75 & 43.82 & 0.16 \\
        DeepScaler & 20.73 & 20.82 & 0.43 \\
        \bottomrule
    \end{tabularx}
\end{table}

\begin{table}[h]
    \captionsetup{justification=raggedright,singlelinecheck=false}
    \centering
    \caption{OPG on additional reasoning domains.}
    \label{tab:domain_opg}
    \vspace{2mm}
    \small
    \begin{tabularx}{\linewidth}{X X c c c}
        \toprule
        \textbf{Model} & \textbf{Domain} & \textbf{Train} & \textbf{Oracle} & \textbf{OPG (\%)} \\
        \midrule
        Qwen2.5-7B & HotpotQA & 77.85 & 78.05 & 0.26 \\
        Qwen2.5-7B & MedQA    & 78.60 & 79.00 & 0.51 \\
        Qwen2.5-7B & LogiQA   & 70.33 & 71.11 & 1.10 \\
        Llama-3-8B & HotpotQA & 78.22 & 78.45 & 0.29 \\
        Llama-3-8B & MedQA    & 83.02 & 83.53 & 0.61 \\
        Llama-3-8B & LogiQA   & 76.48 & 77.19 & 0.92 \\
        \bottomrule
    \end{tabularx}
\end{table}

\subsection{Sensitivity to KL Coefficients}
\label{app:kl_robustness}

We vary the KL coefficient to examine whether the near-zero OPG trend depends on a particular regularization strength. Table~\ref{tab:kl_opg} shows that OPG remains small across all tested KL settings.

\begin{table}[h]
    \captionsetup{justification=raggedright,singlelinecheck=false}
    \centering
    \caption{OPG under different KL settings.}
    \label{tab:kl_opg}
    \vspace{2mm}
    \small
    \begin{tabularx}{\linewidth}{X c c c c}
        \toprule
        \textbf{Dataset} & \textbf{Setting} & \textbf{Train} & \textbf{Oracle} & \textbf{OPG (\%)} \\
        \midrule
        GSM8K & $1.0\times10^{-3}$ & 90.58 & 91.33 & 0.82 \\
        GSM8K & $5.0\times10^{-2}$ & 91.24 & 91.47 & 0.25 \\
        MATH  & $1.0\times10^{-3}$ & 73.68 & 74.86 & 1.58 \\
        MATH  & $5.0\times10^{-2}$ & 73.65 & 73.86 & 0.28 \\
        \bottomrule
    \end{tabularx}
\end{table}

\subsection{Sensitivity to Decoding Parameters}
\label{app:decode_robustness}

We further vary decoding parameters, including temperature and top-$p$, to test whether the near-zero OPG trend is sensitive to the choice of inference configuration. Table~\ref{tab:decode_opg} shows that OPG remains small across all tested settings, suggesting that the observed vanishing-gap phenomenon is robust to reasonable sampling variations at inference time.

\begin{table}[!t]
    \captionsetup{justification=raggedright,singlelinecheck=false}
    \centering
    \caption{OPG under different decoding settings.}
    \label{tab:decode_opg}
    \vspace{2mm}
    \small
    \begin{tabularx}{\linewidth}{X X c c c}
        \toprule
        \textbf{Dataset} & \textbf{Temperature} & \textbf{Train} & \textbf{Oracle} & \textbf{OPG (\%)} \\
        \midrule
        GSM8K & 0.7 & 91.02 & 91.33 & 0.34 \\
        GSM8K & 0.9 & 91.74 & 92.07 & 0.36 \\
        MATH  & 0.7 & 74.05 & 75.06 & 1.35 \\
        MATH  & 0.9 & 74.08 & 75.10 & 1.36 \\
        \midrule
        \textbf{Dataset} & \textbf{Top-$p$} & \textbf{Train} & \textbf{Oracle} & \textbf{OPG (\%)} \\
        \midrule
        GSM8K & 0.7 & 88.52 & 88.76 & 0.27 \\
        GSM8K & 0.9 & 89.22 & 89.49 & 0.30 \\
        MATH  & 0.7 & 69.85 & 70.60 & 1.06 \\
        MATH  & 0.9 & 70.91 & 71.72 & 1.14 \\
        \bottomrule
    \end{tabularx}
\end{table}
\vspace{-2mm}

\end{document}